# CHALLENGE-DEVICE-SYNTHESIS: A MULTI-DISCIPLINARY APPROACH FOR THE DEVELOPMENT OF SOCIAL INNOVATION COMPETENCES FOR STUDENTS OF ARTIFICIAL INTELLIGENCE

**Matías Bilkis[1,2], Joan Moya Kohler[3], Fernando Vilariño[1,2]**

[1]Department of Computer Science, Universitat Autónoma de Barcelona (UAB), (SPAIN)
[2]Computer Vision Center (CVC), Barcelona, (SPAIN)
[3]Department of Social Psychology, Universitat Autónoma de Barcelona (UAB), (SPAIN)

**ABSTRACT**

The advent of Artificial Intelligence is expected to imply profound changes in the short-term. It is therefore imperative for Academia, and particularly for the Computer Science scope, to develop cross-disciplinary tools that bond AI developments to their social dimension. To this aim, we introduce the Challenge-Device-Synthesis methodology (CDS), in which a specific challenge is presented to the students of AI, who are required to develop a device as a solution for the challenge. The device becomes the object of study for the different dimensions of social transformation, and the conclusions addressed by the students during the discussion around the device are presented in a synthesis piece in the shape of a 10-page scientific paper. The latter is evaluated taking into account both the depth of analysis and the level to which it genuinely reflects the social transformations associated with the proposed AI-based device. We provide data obtained during the pilot for the implementation phase of CDS within the subject of Social Innovation, a 6-ECTS subject from the 6th semester of the Degree of Artificial Intelligence, UAB-Barcelona. We provide details on temporalisation, task distribution, methodological tools used and assessment delivery procedure, as well as qualitative analysis of the results obtained.

## 1. INTRODUCTION

### 1.1 Context

The unprecedentedly strong celerity at which the Science & Technology field is currently moving, and particularly Artificial Intelligence (AI), allows for little room to analyse the social implications of novel discoveries and applications. In turn, the segmentation and specialisation in Science **[1,2]**, along with an exponentially-growing AI research community **[3,4]**, indicates the strong necessity to develop new tools to overall address, analyse, and synthesise the potential social impacts.

To this aim, we develop the Challenge-Device-Synthesis (CDS) methodology, conceived as a pedagogical tool to address the development of critical views on the social impact of AI within AI engineering students at degree-course level.

### 1.2 The needs

In this context, it is of utmost importance to reconsider the role that University plays as an interdisciplinary-knowledge agglutinator that tackles the aforementioned problem. In turn, we believe that the format in which knowledge is spread shall, at least, be re-considered, in order to update it to our rapidly changing society. In this regard, connection to societal challenges and alignment with the Sustainable Development Goals **[5]** is an essential and transversal feature that modern teaching modalities tend to incorporate. Here, global challenges posed by the broad scope of the scientific community are furthermore required to be projected towards specific and local challenges, with concrete and situated solutions in our territories.

Our approach is thought to serve as a bridge between the unavoidable growth of scientific knowledge, and the development of specific methodological tools and instruments that jointly allow for collective thinking *within* the learning process, and the validation in concrete scenarios related to our local territory,

*e.g.* Barcelona city. Thus, also valuing the situated knowledge of stakeholders and identifying future users of technological innovation.

An agile methodology **[6]** is not only crucial for the incorporation of technological AI innovations, but also to embrace the expertise from other areas of knowledge, such as psychology, sociology, pedagogy, and philosophy, but also from these less considered but essential knowledges for the proper development and implementation of technology, such as that of the concerned lay people. However, synergies between such distant communities cannot be granted without a solid framework, and the lack of it currently stands as one of the most important challenges in order to guarantee the success of transdisciplinary projects. Here, we humbly propose a prototype of such a potential framework from an unconventional teaching perspective.

### 1.3 Goals

Our objective is to develop a disruptive teaching modality that allows for a transdisciplinary approach to the implication of ground-breaking AI discoveries to their social dimension. From such a main pillar, several objectives can be disentangled, namely: *1)* sparking the collective design of a project prototype, which we define as *the Device,* 2) bonding the device to its social dimension by the evaluating its impact in related communities, a process which is thought to be informed in a ten-pages-length research *paper*, and *3)* identifying links to local communities in territory so to foresee potential follow-up projects of the aforementioned outcomes.

### 1.4 Academic context for the application of the proposed action

The proposed methodological approach was deployed within the teaching context of the Degree of Artificial Intelligence of the Universitat Autònoma de Barcelona (UAB), with students from the 6 ECTS subject of Social Innovation, taking place during the 6th degree semester, in 2024. The lectures were divided into 2 weekly sessions of 2 hours each, for 15 weeks. The first session was designed to be oriented fundamentally to the presentation and discussion of new theoretical contents, while the second session was designed to the development of the reflections around the device and the integration of the results into the paper.

The subject of *Social Innovation* integrates the challenge-based approach, including challenge-based learning **[7]**, as methodological contributions to teaching, in a context in which UAB is a member of the ECIU European University **[8]**, which is aiming at the identification of actual challenges from the territory, to be mapped and translated to specific actions within the learning context.

The teaching staff was from two different departments of the UAB, namely: 1) the Computer Science Department, which is the coordinator of different degrees and master's degrees in computer science at the School of Engineering, and 2) the Social Psychology Department, which coordinates different degrees in the Faculty of Humanities and Social Sciences. Both Departments contribute from different perspectives to the discussion of social impacts and acceptability of the technology, potential social changes of the created AI tools, and prospective identification of new habits, regulatory learning, etc. It also provides a solid theoretical background for the responsible research and the critical appraisal of the impacts brought by technology.

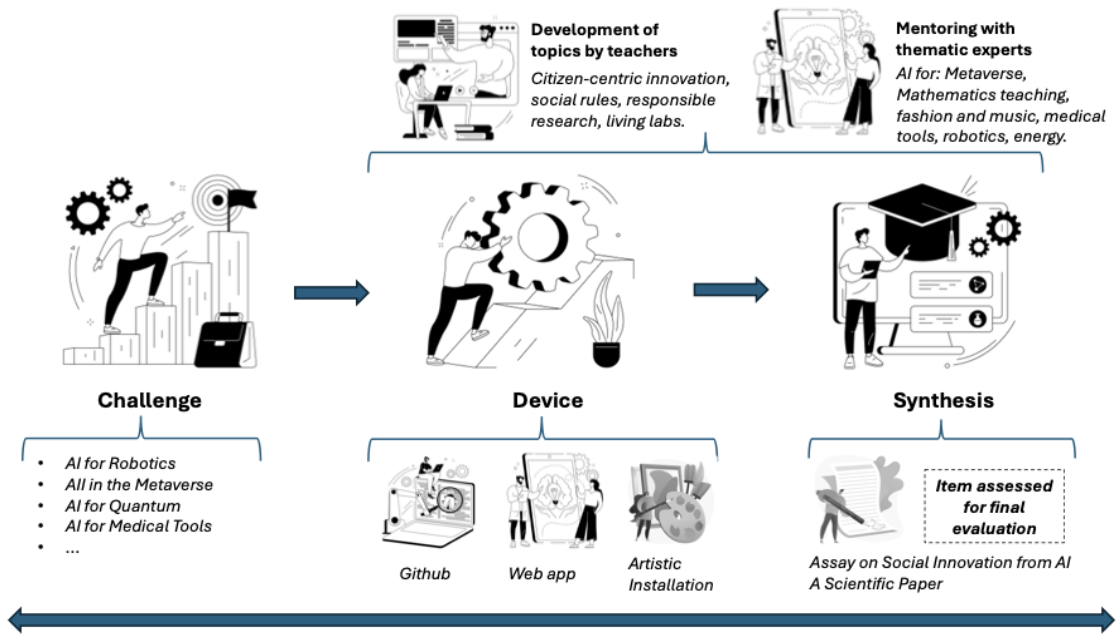

*Figure 1: Graphical approach to the Challenge-Device-Synthesis. In bold, the general components. In cursive, the specific adaptations to the subject Social Innovation on the AI degree.*

## 2. METHODOLOGY

The Challenge-Device-Synthesis methodology (CDS) is thought to be carried out by an interdisciplinary teaching team (which in the case reported in this paper is comprised by social psychologist, a theoretical physicist and a computer scientist). Such a background heterogeneity crucially warrants the diversity both in approaching the social impact of AI from a wide variety of angles, but also to enrich the quality of the feedback provided to the students. The methodological body can be divided in 3 parts:

1) Challenge identification and adoption,
2) device construction, and
3) development of the synthesis.

Figure 1 provides a graphical representation of the methodology proposed, with all its general components, and the specific implementation within the academic context of "Social Innovation".

### 2.1 Challenge identification and adoption

Students are provided with a list of *Challenges* by the teaching team, a list of pre-identified real-world scenarios that might radically be affected by the advent of AI. Each of the Challenges appearing in the list is aimed to be a field in AI research, having strong roots with society. Within the specific academic context, care must be taken when defining such a list; since the targeted alumni was heavily skilled in computer programming and had little to null experience in social and human research, the list of Challenges should be sufficiently interesting so as to guarantee strong motivation of the alumni across the whole course. On the other hand, the social link of each Challenge should be evident enough to facilitate its social-impact study. For this pilot, the Challenges list was composed of the fields of "Health, Robotics, Generative technologies, and Quantum Machine Learning".

### 2.2 Device construction

The alumni are split into groups of 3-6 students, and each group is asked to pick one Challenge among the list. Departing from such a field, the group is requested to conceive a *Device.* The latter constitutes a concrete application of an AI technology related to the Challenge, and is thought to be implemented by the students as a fully-working prototype by the end of the course. In practice, the Device plays two important roles: On the one hand, it serves a route that links the Challenge with the specific societal impacts to be studied, rather than an object for evaluation in the subject. On the other hand, it serves as a lighthouse for students, in order to keep the motivation spark alive as the course evolves. For our specific academic context where students are highly skilled in AI programming, it is relevant to recall that students might not *a priori* understand the importance of analysing social counterparts. Examples of Devices are (but not restricted to): Github sites (open source repositories), artistic creations, machine learning libraries or business models.

**2.3 Development of the synthesis**

Finally, the synthesis instrument -which bonds the AI technology (as captured by the Device development) to its social dimension- consists of an academic assay in the shape of a scientific paper. This paper hinges on the device as a source of deflection for analysing and discussing the potential societal transformations created by it. Importantly, it is the *only* piece of work which will be finally evaluated by the teachers. The latter is evaluated taking into account both the depth of analysis and the level to which it genuinely reflects the social transformations associated with the proposed AI-based device. The Device is implicitly evaluated within the paper in terms of its relevance to deploy the social innovation dimensions.

**2.4 Other methodological components**

The general framework is endowed with the following methodological components providing evidences and points of discussion:

- **Context and Practice lectures.** Lectures can be classified into *Context* and *Practice*. *Context* lectures are aimed to provide tools from Social & Human sciences in order to enrich the scope of the final paper, while *Practice* lectures serve as a space to develop the device and spark the discussion among the students on the different topics shared in the Context lectures. Some agility is required for the teaching staff to optimally manage time and resource allocation to boost the project's progress through both technical and non technical aspects. In the specific academic context, the weekly distribution of topics for the Context lectures is presented in Figure 2.
- **Interviews with experts.** In order to steer the social impact analysis towards the respective communities, a series of interviews are scheduled by the mid-term of the course. Experts are invited to attend the lectures and act as mentors during 2 hours for the discussion of the device and the potential innovative solutions. Such an expert is contacted by the teaching staff, and a rigorous interview plan is discussed in full detail beforehand[1]. The goal of the expert-interview is to provide a perspective on the maturity and overall potential of Device, from a highly-technical point of view.
- **Interviews with non-experts.** Each group is required to find a *lay person*, *e.g.* a non-expert individual that is however affected by the Challenge. As an example, if the Challenge was related to Health, such a lay person can be a nurse or a doctor, or a patient. Moreover, the group is not only in charge of reaching out to the individual, but also to schedule the interview in full detail. The overall goal of such an interview is to enrich the development of the challenge from a social (though non-technical) perspective.

---

[1] As a case of illustrative example, and in the specific academic context of the UAB, if the Challenge was related to Health, the expert could have been chosen as the lead developer of the public HealthCare Virtual Assistant. Particularly in Catalonia, the main app for the Public Health System is La MevaSalut **[9]**, and one good candidate for an expert would be someone related to its actual technical deployment.

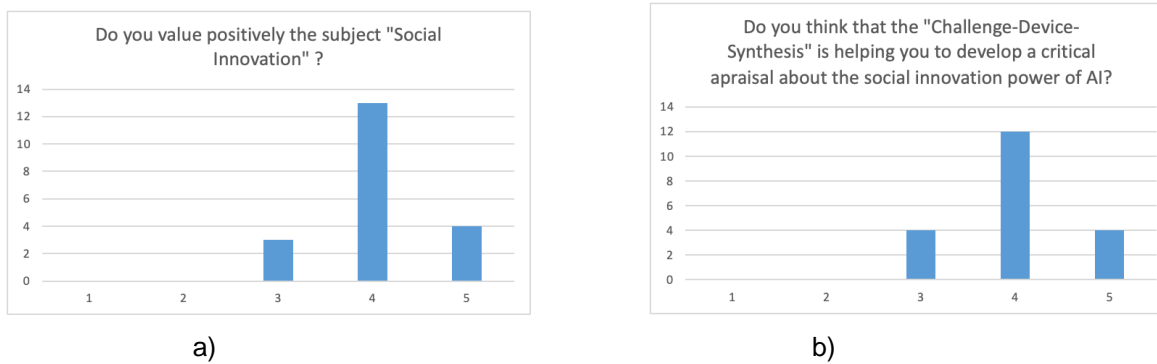

*Figure 2: Specific calendarization framework of the course for the Context lectures, together with the check-points and follow-up sessions for the Device construction.*

## 3. RESULTS

Preliminary results yielded by the pilot, using quantitative and qualitative data obtained from direct anonymous surveys on 20 students, provide us with positive assessment from the perspective of the student. Qualitative analysis of the devices produced provides us with a rich diversity of devices that allow for an understanding of the actual transformative impact that the AI can have at societal level. These results also provide evidence about the natural acquisition of the social competences needed to be integrated in the curriculum of the future AI engineers.

### 3.1 Results from the survey

a)                                              b)

*Figure 3: Results on perception from the students on the value (a) and the development of critical appraisal (b) of the methodological framework used.*

With regard to the perception of the students on the subject implementing the methodological framework proposed, and as presented in Figure 3 (a), the students acknowledged with a 85% of consensus (17/20) that they found either valuable (12) or very valuable (5) the subject. Regarding the methodological framework (see Figure 3 (b)), 80% (16/20) found the approach helpful or very helpful to develop a critical appraisal about the social innovation power of AI. It is relevant to identify that from the valid responses no response indicated no value or reduced value for both questions.

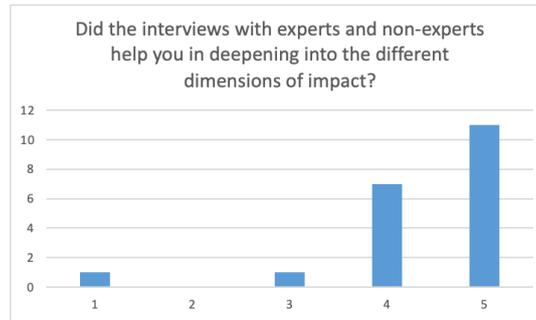

*Figure 4: Identification of the interviews as relevant methodological instruments.*

Regarding the relevance of the interviews with the experts and non-experts, 90% (18/20) identifying as helpful or very helpful the interviews. It is very relevant to point out here that there is a clear majority of students providing the maximum calcification.

### 3.2 Qualitative analysis of the devices in the context of the challenges

In the following we summarise a list of specific examples of the Challenges considered in the current course; we note that such a list is not exhaustive and that there are more groups actively working on different Challenges. For each of the projects we specify potential social impacts (as identified by the corresponding group of students), the device, and corresponding interviewees.

- ***Dressing by AI:*** The Challenge identified touches on the Generative AI, with strong links to the metaverse. Here, the AI-tech impact is identified in the artistic community performing in the metaverse, with a lack of concrete tools to develop a virtual identity.
- ***OpenMath***: The challenge identified hinges on Generative AI, with a strong emphasis in Large Language Models (LLMs) for pedagogical purposes. Here, the social target is constituted by high-school maths courses, with the aim of personalising the learning experience in order to help both the teacher and the alumni during the course.
- ***Storia:*** The challenge here relates to Health, and the project aims to tackle loneliness felt by hospitalised children which in turn need to be isolated due to their medical conditions. Here, the social impact touches hospital communities (mainly young patients, but also doctors), and the device consists in developing a software that overall allows hospitalised children to virtually interact by jointly creating a tale. Importantly, such a tale is generated by sketches and prompts given by the children.
- ***Quantum energy:*** The challenge here hinges on the advent of Quantum Technologies & Quantum Machine Learning. The social impact is related to the energy consumption of AI technologies and potential savings that quantum phenomena (e.g. quantum correlations between system and reservoir) might allow.
- ***WaterMarkingAI:*** The challenge falls in the Generative machine learning, and touches copyright issues by the usage of LLMs. Here, the device consists in developing a watermark for text generated out of an AI model (such as ChatGPT). The social impact has many actors, among them students at different levels, but also companies which can potentially be required to acknowledge the automatic text generation by law.

For all the challenges identified, the proposed devices tackle effective and measurable future impacts, and relate to social innovations that will eventually happen provided those devices deploy their particular potential.

### 4. DISCUSSION & FUTURE WORK

We have demonstrated at the outset the challenge that modern societies face in integrating different types and levels of knowledge, especially important when attempting to incorporate technologies with as significant change potentials as AI **[11]**. To meet this challenge, we have proposed a study program, "Challenge-Device-Synthesis" (CDS), implemented at the Universitat Autònoma de Barcelona. This work illustrates how the integration of technical, social, historical, and practical or situated knowledge not only enriches the training of future AI engineers but also promotes a deeper understanding of the social implications of technology. This multidisciplinary approach allows students to not only build technically effective devices but also to consider how these devices can effectively serve society and adapt to various social realities.

Through the process, students manage to incorporate social concepts and approaches, whose handling allows them to learn about the social reality in which their future device will be immersed. For this purpose, they acquire competencies related to social sciences, techniques to understand social reality, and concepts from Science and Technology Studies that allow them to construct a critical view of the effects and potentialities of their devices.

The final synthesis, in the form of a scientific article, reflects not only the technical impact of the devices created but also their potential to transform society in significant and sustainable ways. The integration of interviews with experts and laypeople in the development process underlines the importance of incorporating multiple perspectives and ensures that the proposed solutions have real relevance and applicability **[11]**.

In a world increasingly dependent on complex technologies, engineers must be more than mere technicians; the need for sensitivity and interdisciplinary understanding in the development of AI is more crucial than ever to generate common good **[12]**. This integrative approach is not only essential for the development of technologies that are sustainable and beneficial for all, but also prepares students to be innovative and responsible in the field of AI, equipped to face future challenges with a broad and comprehensive vision.

In this sense, the Challenge-Device-Synthesis (CDS) methodology has functioned as a sort of space for harpedonaptia, following philosopher Michel Serres **[13]**. Serres' Harpedonapta, an ancient Egyptian figure dedicated to precise land measurement, serves as an eloquent metaphor to highlight the importance of a multidisciplinary vision in engineering and the development of advanced technologies like artificial intelligence (AI). Just as the Harpedonapta integrated knowledge of geometry, history, law and astronomy to fulfil his duties, modern engineers must incorporate a wide range of disciplines to design technological solutions that are truly effective and socially responsible.

From here, this harpedonaptic, integrative proposal presents some promising directions that can be explored to continue enriching the training of students in artificial intelligence. Firstly, it opens the door to expanding the list of challenges and adapting to new emerging and urgent themes such as climate change, urban sustainability, or equity in technological access. This could involve more academic departments and experts from fields such as ecology, urban planning, and social studies.

In the same vein, work can be approached that allows establishing collaborations with universities in different parts of the world to enable students to work on challenges that are particular to different geographical and cultural contexts.

This project opens the door to creating forms of continuity, so that the projects developed in the course can be continued, not only in future semesters but also in real situations through the possibility of presenting these projects during Universitat Autònoma de Barcelona's innovation week.

Finally, the project also opens the door to finding even more potent ways to foster greater local community participation in the device development process, from the ideation phase to evaluation, ensuring that the developed solutions are inclusive and meet the needs of the end-users.

## 5. ACKNOWLEDGEMENTS

This work was done partially under the support of the project Cátedra UAB-Cruïlla, funded by the Spanish Government (TSI-100929-2023-2 (Prov)). All our acknowledgement to the experts participating in the interviews and mentoring. All our acknowledgement to the students, the actual protagonist of the future social impacts of AI.